\documentclass[10pt,journal,compsoc]{IEEEtran}

\usepackage[nocompress]{cite}
\usepackage[pdftex]{graphicx}
\graphicspath{figs}
\DeclareGraphicsExtensions{.pdf}
\usepackage{array}
\usepackage[caption=false,font=footnotesize,labelfont=sf,textfont=sf]{subfig}
\usepackage{fixltx2e}
\usepackage{dblfloatfix}
\let\MYoriglatexcaption\caption
\renewcommand{\caption}[2][\relax]{\MYoriglatexcaption[#2]{#2}}
\usepackage{url}
\usepackage{booktabs,multirow}
\usepackage[ruled]{algorithm2e}

\usepackage{amsmath,amsfonts,bm}

\def\eqref#1{equation~\ref{#1}}

\def\1{\bm{1}}

\def\vtheta{{\bm{\theta}}}
\def\vgamma{{\bm{\gamma}}}
\def\vbeta{{\bm{\beta}}}
\def\va{{\bm{a}}}

\def\vs{{\bm{s}}}
\def\vt{{\bm{t}}}

\def\vv{{\bm{v}}}
\def\vw{{\bm{w}}}
\def\vx{{\bm{x}}}
\def\vy{{\bm{y}}}

\def\mA{{\bm{A}}}

\def\mF{{\bm{F}}}
\def\mG{{\bm{G}}}

\def\mI{{\bm{I}}}

\def\mM{{\bm{M}}}

\def\mW{{\bm{W}}}

\DeclareMathAlphabet{\mathsfit}{\encodingdefault}{\sfdefault}{m}{sl}
\SetMathAlphabet{\mathsfit}{bold}{\encodingdefault}{\sfdefault}{bx}{n}

\def\gA{{\mathcal{A}}}

\def\gL{{\mathcal{L}}}

\def\gS{{\mathcal{S}}}

\newcommand{\param}{\vtheta}

\newcommand{\E}{\mathbb{E}}

\newcommand{\R}{\mathbb{R}}
\newcommand{\T}{\top}

\def\etal{\textit{et al.}}
\def\eg{\textit{e.g}}

\usepackage{xcolor}

\begin{document}
\title{
Scalable and Practical Natural Gradient \\ 
for Large-Scale Deep Learning
}

\author{Kazuki~Osawa,~\IEEEmembership{Student~Member,~IEEE,}
       Yohei~Tsuji,
       Yuichiro~Ueno,
       Akira~Naruse, \\
       Chuan-Sheng~Foo,
       and~Rio~Yokota
\IEEEcompsocitemizethanks{
\IEEEcompsocthanksitem 
K. Osawa and Y. Tsuji are with the Tokyo Institute of Technology, Japan.\protect\\
E-mail: (oosawa.k.ad, tsuji.y.ae)@m.titech.ac.jp
\IEEEcompsocthanksitem 
Y. Ueno and R. Yokota are with the Tokyo Institute of Technology, Japan and
AIST-Tokyo Tech RWBC-OIL, AIST, Japan.\protect\\
E-mail: ueno.y.ai@m.titech.ac.jp, rioyokota@gsic.titech.ac.jp
\IEEEcompsocthanksitem 
C.S. Foo is with the Institute for Infocomm Research, A*STAR, Singapore.\protect\\
E-mail: foo\_chuan\_sheng@i2r.a-star.edu.sg
\IEEEcompsocthanksitem 
A. Naruse is with NVIDIA, Japan.\protect\\
E-mail: anaruse@nvidia.com
}
\thanks{Manuscript received MONTH DD, 20XX; revised MONTH DD, 20XX.}}
\IEEEtitleabstractindextext{%
\begin{abstract}
Large-scale distributed training of deep neural networks 
results in models with worse generalization performance as a result of the increase in the effective mini-batch size.
Previous approaches attempt to address this problem by varying the learning rate 
and batch size over epochs and layers, 
or \textit{ad hoc} modifications of batch normalization. 
We propose \textit{Scalable and Practical Natural Gradient Descent} (SP-NGD),
a principled approach for training models that allows them to attain similar generalization performance to models trained with first-order optimization methods, 
but with accelerated convergence. Furthermore, SP-NGD scales to large mini-batch sizes with a negligible computational overhead as compared to first-order methods. 
We evaluated SP-NGD on a benchmark task
where highly optimized first-order methods are available as references: training a ResNet-50 model for image classification on ImageNet. We demonstrate convergence to a top-1 validation accuracy of 75.4\% in 5.5 minutes using a mini-batch size of 32,768 with 1,024 GPUs, as well as an accuracy of 74.9\% with an extremely large mini-batch size of 131,072 in 873 steps of SP-NGD.

\end{abstract}

}

\maketitle

\IEEEdisplaynontitleabstractindextext

 \ifCLASSOPTIONpeerreview
 \begin{center} \bfseries EDICS Category: 3-BBND \end{center}
 \fi
\IEEEpeerreviewmaketitle

\IEEEraisesectionheading{\section{Introduction}\label{sec:introduction}}

\IEEEPARstart{A}{s} the size of deep neural network models and the data which they are trained on continues to increase rapidly,
the demand for distributed parallel computing is increasing.
A common approach for achieving distributed parallelism in deep learning is to use the data-parallel approach,
where the data is distributed across different processes while the model is replicated across them.
When the mini-batch size per process is kept constant to increase the ratio of computation over communication,
the effective mini-batch size over the entire system grows proportional to the number of processes.

When the mini-batch size is increased beyond a certain point, the generalization performance starts to degrade.
This generalization gap caused by large mini-batch sizes have been studied extensively
for various models and datasets \cite{shallue2019}.
Hoffer \etal \cite{hoffer2018} attribute this generalization gap to the limited number of updates, and suggest to train longer.
This has lead to strategies such as scaling the learning rate proportional to the mini-batch size,
while using the first few epochs to gradually warmup the learning rate \cite{smith2018a}.
Such methods have enabled the training for mini-batch sizes of 8K,
where ImageNet \cite{deng2009} with ResNet-50 \cite{he2016a} could be trained for 90 epochs
to achieve 76.3\% top-1 validation accuracy in 60 minutes \cite{goyal2017}.
Combining this learning rate scaling with other techniques such as RMSprop warm-up,
batch normalization without moving averages, and a slow-start learning rate schedule,
Akiba \etal \cite{akiba2017} were able to train the same dataset and model with a mini-batch size of 32K
to achieve 74.9\% accuracy in 15 minutes.

More complex approaches for manipulating the learning rate were proposed, such as LARS \cite{you2017},
where a different learning rate is used for each layer by normalizing them with
the ratio between the layer-wise norms of the weights and gradients.
This enabled the training with a mini-batch size of 32K
without the use of \textit{ad hoc} modifications, which achieved 74.9\% accuracy
in 14 minutes (64 epochs) \cite{you2017}.
It has been reported that combining LARS with counter intuitive modifications to the batch normalization,
can yield 75.8\% accuracy even for a mini-batch size of 65K \cite{jia2018}.

The use of small batch sizes to encourage rapid convergence in early epochs,
and then progressively increasing the batch size is yet another successful approach.
Using such an adaptive batch size method, Mikami \etal \cite{mikami2018} were able to train in 122 seconds
 with an accuracy of 75.3\%,
 and Yamazaki \etal \cite{yamazaki2019} were able to train in 75 seconds with a accuracy of 75.1\%.
The hierarchical synchronization of mini-batches have also been proposed \cite{lin2018},
but such methods have not been tested at scale to the extent of the authors' knowledge.

\begin{table*}[t]
    \centering
    \caption{Training time and top-1 single-crop validation accuracy of ResNet-50 for ImageNet reported by related work and this work.}
    \begin{tabular}{c c c r c r c c c}
        \toprule
        &\bf Hardware&\bf Software&\bf Batch size&\bf Optimizer&\bf \# Steps& \bf Time/step & \bf Time & \bf Accuracy\\
        \midrule
        Goyal \etal \cite{goyal2017} & Tesla P100 $\times$ 256 & Caffe2 & 8,192 & SGD & 14,076 & 0.255~s & 1 hr & {\bf 76.3}~\% \\
        You \etal \cite{you2017} & KNL $\times$ 2048 & Intel Caffe & 32,768 & SGD & 3,519 & 0.341~s & 20 min & 75.4~\% \\
        Akiba \etal \cite{akiba2017} & Tesla P100 $\times$ 1024 & Chainer & 32,768 & RMSprop/SGD & 3,519 & 0.255~s & 15 min & 74.9~\% \\
        You \etal \cite{you2017} & KNL $\times$ 2048 & Intel Caffe & 32,768 & SGD & 2,503 & 0.335~s & 14 min & 74.9~\% \\
        Jia \etal \cite{jia2018} & Tesla P40 $\times$ 2048 & TensorFlow & 65,536 & SGD & 1,800 & 0.220~s & 6.6 min & 75.8~\% \\
        Ying \etal \cite{ying2018} & TPU v3 $\times$ 1024 & TensorFlow & 32,768 & SGD & 3,519 & {\bf 0.037~s} & 2.2 min & {\bf 76.3}~\% \\ 
        Mikami \etal \cite{mikami2018} & Tesla V100 $\times$ 3456 & NNL & 55,296 & SGD & 2,086 & 0.057~s & 2.0 min & 75.3~\% \\
        Yamazaki \etal \cite{yamazaki2019} & Tesla V100 $\times$ 2048 & MXNet & 81,920 & SGD & 1,440 & 0.050~s & {\bf 1.2 min} & 75.1~\% \\
        \midrule
        \multirow{6}{*}{This work} & Tesla V100 $\times$ 128 & \multirow{6}{*}{Chainer} & 4,096 & \multirow{6}{*}{SP-NGD} & 10,948 & 0.178~s & 32.5 min & 74.8~\% \\
        & Tesla V100 $\times$ 256 & & 8,192 & & 5,434 & 0.186~s & 16.9 min & 75.3~\% \\
        & Tesla V100 $\times$ 512 & & 16,384 & & 2,737 & 0.149~s & 6.8 min & 75.2~\% \\
        & Tesla V100 $\times$ 1024 & & 32,768 & & 1,760 & 0.187~s & {\bf 5.5 min} & 75.4~\% \\
        & n/a  & & 65,536 & & 1,173 & n/a & n/a & 75.6~\% \\
        & n/a  & & {\bf 131,072} & & {\bf 873} & n/a & n/a & 74.9~\% \\
        \bottomrule
    \end{tabular}
    \label{tab:compare}
\end{table*}

\begin{figure*}[t]
    \centering
    \includegraphics[width={\textwidth}]{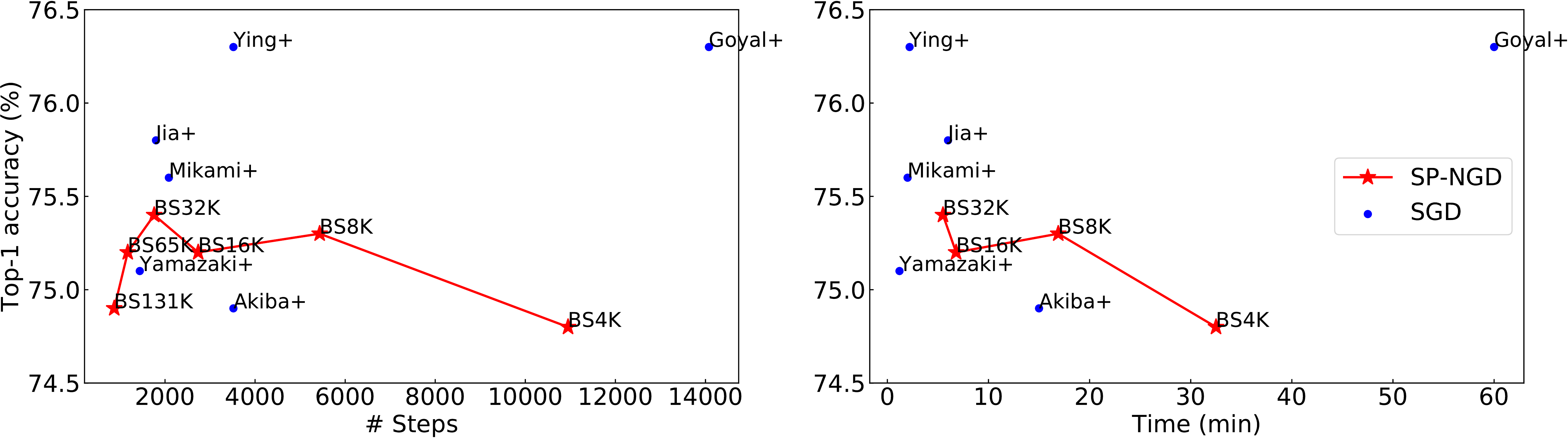}
    \caption{Top-1 validation accuracy vs the number of steps to converge (left) 
    and vs training time (right) 
    of ResNet-50 on ImageNet (1000 class) classification
    by related work with SGD and this work with \textit{Scalable and Practical NGD} (SP-NGD).}
    \label{fig:accuracy_iter}
\end{figure*}

In this work, we take a more mathematically rigorous approach 
to close the generalization gap when large mini-batches are used.
Our approach builds on Natural Gradient Descent (NGD) \cite{amari1998},
a second-order optimization method that leverages curvature information to accelerate optimization.
This approach is made feasible by the use of large mini-batches, which enables stable estimation of curvature even in models with a large number of parameters. To this end, we propose an efficient distributed NGD design that scales to massively parallel settings and large mini-batch sizes. In particular, we demonstrate scalability to batch sizes of 32,768 over 1024 GPUs across 256 nodes. 
Another unique aspect of our approach is the accuracy 
at which we can approximate the Fisher information matrix (FIM) 
when compared to other second-order optimization methods. 
Unlike methods that use very crude approximations of the FIM, such as the TONGA \cite{roux2008},
Hessian free methods \cite{martens2010},
we adopt the Kronecker-Factored Approximate Curvature (K-FAC) method \cite{martens2015}.
The two main characteristics of K-FAC are that it converges faster than first-order
stochastic gradient descent (SGD) methods,
and that it can tolerate relatively large mini-batch sizes without any \textit{ad hoc} modifications.
K-FAC has been successfully applied to convolutional neural networks \cite{grosse2016},
distributed memory training of ImageNet \cite{ba2017}, recurrent neural networks \cite{martens2018},
Bayesian deep learning \cite{zhang2018a}, reinforcement learning \cite{wu2017}
, and Transformer models \cite{zhang2019a}.

Our contributions are:
\begin{itemize}
    \item {\bf Extremely large mini-batch training.}
    	We were able to show for the first time 
    	that approximated NGD can achieve similar generalization capability
    	compared to highly optimized SGD, by training ResNet-50 on ImageNet classification as a benchmark.
	    We converged to over 75\% top-1 validation accuracy for large mini-batch sizes of 4,096, 8,192, 16,384, 32,768 and 65,536.
        We also achieved 74.9\% with an extremely large mini-batch size of 131,072, which took only 873 steps.
    \item {\bf Scalable natural gradient.}
        We propose a distributed NGD design using data and model hybrid parallelism 
        that shows \textit{superlinear} scaling up to 64 GPUs.
    \item {\bf Practical natural gradient.}
        We propose practical NGD techniques based on analysis of 
        the FIM estimation in large mini-batch settings.
        Our practical techniques make the overhead of NGD compare to SGD almost negligible.
        Combining these techniques with our distributed NGD,
        we see an ideal scaling up to 1024 GPUs as shown in Figure~\ref{fig:secpiter}.
    \item {\bf Training ResNet-50 on ImageNet in 5.5 minutes.}
        Using 1024 NVIDIA Tesla V100, we achieve 75.4 \% top-1 accuracy with ResNet-50 for ImageNet in 5.5 minutes 
        (1760 steps = 45 epochs, including a validation after each epoch).
        The comparison is shown in Figure~\ref{fig:accuracy_iter} and Table~\ref{tab:compare}.
\end{itemize}
A preliminary version of this manuscript was published previously \cite{osawa2019a}.  
Since then, the performance optimization of the distributed second-order optimization has been studied \cite{tsuji2019},
and our distributed NGD framework has been applied to accelerate Bayesian deep learning 
with the natural gradient at ImageNet scale \cite{osawa2019}.
We extend the previous work and 
propose \textit{Scalable and Practical Natural Gradient Descent} (SP-NGD) framework, 
which includes more detailed analysis on the FIM estimation and 
significant improvements 
on the performance of the distributed NGD.

\section{Related work}

With respect to large-scale distributed training of deep neural networks,
there have been very few studies that use second-order methods.
At a smaller scale, there have been previous studies that used K-FAC to train ResNet-50 on ImageNet \cite{ba2017}.
However, the SGD they used as reference was not showing state-of-the-art Top-1 validation accuracy
(only around 70\%), so the advantage of K-FAC over SGD that they claim was not obvious from the results.
In the present work, we compare the Top-1 validation accuracy with state-of-the-art SGD methods for large mini-batches
mentioned in the introduction (Table~\ref{tab:compare}).

The previous studies that used K-FAC to train ResNet-50 on ImageNet \cite{ba2017}
also were not considering large mini-batches and were only training with mini-batch size of 512 on 8 GPUs.
In contrast, the present work uses mini-batch sizes up to 131,072, which is equivalent to 32 per GPU on 4096 GPUs,
and we are able to achieve a much higher Top-1 validation accuracy of 74.9\%.
Note that such large mini-batch sizes can also be achieved by accumulating the gradient over multiple iterations
before updating the parameters, which can mimic the behavior of the execution on many GPUs
without actually running them on many GPUs.

The previous studies using K-FAC also suffered from large overhead of the communication
since they used a parameter-server approach for their TensorFlow \cite{abadi2016} 
implementation of K-FAC with a single parameter-server.
Since the parameter server requires all workers to
send the gradients and receive the latest model's parameters from the parameter server,
the parameter server becomes a huge communication bottleneck especially at large scale.
Our implementation uses a decentralized approach using MPI/NCCL\footnote{https://developer.nvidia.com/nccl} collective communications among the processes.
The decentralized approach has been used in high performance computing for a long time,
and is known to scale to thousands of GPUs without modification. 
Although, software like Horovod\footnote{https://github.com/horovod/horovod} can
alleviate the problems with parameter servers by working as a TensorFlow wrapper for NCCL, 
a workable realization of K-FAC requires solving many engineering and modeling challenges, and our solution is the first one that succeeds on a large scale task.

\section{Notation and Background}

\subsection{Mini-batch Stochastic Learning}
Throughout this paper, we use $E[\cdot]$ 
to denote the empirical expectation among the samples in 
the mini-batch $\{(\vx,\vt)\}$, 
and compute the {\it cross-entropy loss} for a supervised learning as
\begin{equation}
    \label{eq:loss}
    \mathcal{L}(\boldsymbol\theta) 
    =
    E[-\log p_{\boldsymbol\theta}(\vt|\vx)]\,.
\end{equation}
where $\vx, \vt$ are the training input and label (one-hot vector),
$p_{\param}(\vt|\vx)$ is 
the likelihood of each sample $(\vx,\vt)$ 
calculated by the probabilistic model using a feed-forward deep neural network (DNN) 
with the parameters $\param\in\R^N$.

For the standard mini-batch stochastic gradient descent (SGD), 
the parameters $\param$ are updated based on the gradient of the loss function at the current point:
\begin{align}
    \param^{(t+1)}
    &\leftarrow
    \param^{(t)}
    -
    \eta
    \nabla\mathcal{L}(\param^{(t)})\,.
\end{align}
where $\eta>0$ is the learning rate.

\subsection{Natural Gradient Descent in Deep Learning}
\begin{figure*}[t]
    \centering
    \includegraphics[width=0.8\textwidth]{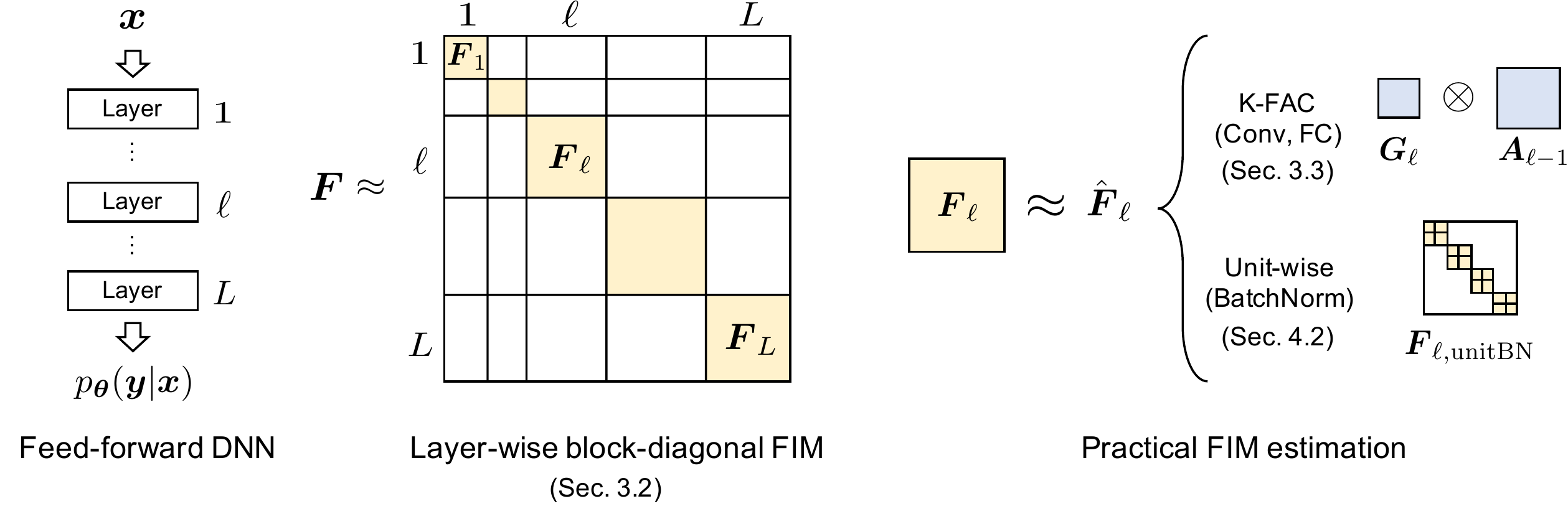}
    \caption{Illustration of Fisher information matrix approximations for feed-forward deep neural networks used in this work.}
    \label{fig:practical_fim}
\end{figure*}
{\it Natural Gradient Descent} (NGD)\cite{amari1998} is an optimizer
which updates the parameters using the first-order gradient of the loss function
preconditioned by the {\it Fisher information matrix} (FIM) of the probabilistic model:
\begin{equation}
    \param^{(t+1)}
    \leftarrow
    \param^{(t)}
    -
    \eta
    \left(
    \mF+\lambda\mI
    \right)^{-1}
    \nabla\mathcal{L}(\param^{(t)})\,.
\end{equation}
The FIM $\mF\in\R^{N\times N}$ of a DNN 
with the learnable parameter $\param\in\mathbb{R}^N$ is defined as:
\begin{equation}
    \mF:=
    \E_{\vx\sim q}\left[
    \E_{\vy\sim p_{\param}}
    \left[
    \nabla\log p_{\param}(\vy|\vx)
    \nabla\log p_{\param}(\vy|\vx)^\T
    \right]
    \right]\,.
\end{equation}
$\E_v[\cdot]$ is an expectation w.r.t. the random variable $v$,
and $q$ is the training data distribution. 
To limit the step size, a {\it damping} value $\lambda>0$ is added to 
the diagonal of $\mF$ before inverting it.
In the training of DNNs, 
the FIM may be thought of as the curvature matrix in parameter space \cite{amari1998,martens2015,botev2017}.

To realize an efficient NGD training procedure for deep neural networks, we make the
following approximations to the FIM: 
\begin{itemize}
    \item {\bf Layer-wise block-diagonal approximation.} 
        We assume that the correlation between parameters in different layers (Figure~\ref{fig:practical_fim}) is negligible and can be ignored.
        This assumption significantly reduces the computational cost of inverting $\mF$
        especially when $N$ is large.
    \item {\bf Stochastic natural gradient.} 
        We approximate the expectation over the input data distribution $\E_{\vx\sim q}[\cdot]$
        using the empirical expectation over a mini-batch $E[\cdot]$.
        This enables estimation of $\mF$ during mini-batch stochastic learning.
    \item {\bf Monte Carlo estimation.} 
        We approximate the expectation over the model predictive distribution 
        $\E_{\vy\sim p_{\param}}[\cdot]$ using a single Monte Carlo sample (a single backward-pass).
        We note that for a $K$-class classification model, $K$ backward-passes are required to approximate $\mF$. 
\end{itemize}
Using these approximations, we estimate the FIM 
$\mF_{\ell}\in\R^{N_\ell\times N_\ell}$ 
for the $\ell$-th layer using a Monte Carlo sample $\vy\sim p_{\param}(\vy|\vx)$ for each input $\vx$ in a mini-batch as
\begin{equation}
    \label{eq:fisher_1mc}
    \mF_{\ell}\approx
    \hat{\mF}_{\ell}:=
    E\left[
    \nabla_{\vw_{\ell}}\log p_{\param}(\vy|\vx)
    \nabla_{\vw_{\ell}}\log p_{\param}(\vy|\vx)^\T
    \right]\,.
\end{equation}
With this $\hat{\mF}_\ell$,
the parameters $\vw_\ell\in\R^{N_\ell}$ for 
the $\ell$-th layer are then updated using the FIM preconditioned gradients: 
\begin{equation}
    \label{eq:ngd}
    \vw_{\ell}^{(t+1)}
    \leftarrow
    \vw_{\ell}^{(t)}
    -
    \eta
    \left(\hat{\mF}^{(t)}_{\ell}+\lambda \mI\right)^{-1}
    \nabla_{\vw_\ell}\mathcal{L}^{(t)}\,.
\end{equation}
Here $\nabla_{\vw_{\ell}}\mathcal{L}^{(t)}\in\R^{N_\ell}$ denotes 
the gradient of the loss function w.r.t. $\vw_\ell$ for $\vw_\ell=\vw_\ell^{(t)}$.

\subsection{K-FAC}
Kronecker-Factored Approximate Curvature (K-FAC) \cite{martens2015} is 
a second-order optimization method for deep neural networks, 
that is based on an accurate and mathematically rigorous approximation of the FIM.
Using K-FAC, we further approximate the FIM $\mF_{\ell}$ for $\ell$-th layer 
as a Kronecker product of two matrices (Figure~\ref{fig:practical_fim}):
\begin{equation}
    \label{eq:fim}
    {\mF}_{\ell}
    \approx
    {\hat{\mF}}_{\ell}
    ={\mG}_{\ell}\otimes{\mA}_{\ell-1}
    \,.
\end{equation}
This is called {\it Kronecker factorization} and 
${\mG}_{\ell},{\mA}_{\ell-1}$ are called {\it Kronecker factors}.
${\mG}_{\ell}$ is computed from the gradient of the loss
w.r.t. the output of the $\ell$-th layer, 
and ${\mA}_{\ell-1}$ 
is computed from the activation of the $(\ell-1)$-th layer 
(the input of $\ell$-th layer). 

The definition and the sizes of 
the Kronecker factors $\mG_{\ell}/\mA_{\ell-1}$ depend on 
the dimension of the output/input and the type of layer 
\cite{martens2015,grosse2016,martens2018,zhang2019a}.

\subsubsection{K-FAC for fully-connected layers.}
In a fully-connected (FC) layer of a feed-forward DNN,
the output $\vs_{\ell}\in\R^{d_\ell}$ is calculated as
\begin{equation}
   \vs_{\ell}
   \leftarrow
   \mW_{\ell}
   \va_{\ell-1},
\end{equation}
where $\va_{\ell-1}\in\R^{d_{\ell-1}}$ is the input to this layer
(the activation from the previous layer),
and $\mW_{\ell}\in\R^{d_{\ell}\times d_{\ell-1}}$ is the weight matrix 
(the bias is ignored for simplicity).
The Kronecker factors for this FC layer are defined as
\begin{equation}
\label{eq:kfac_fc}
\begin{split}
    \mG_{\ell}
    &:=
    E\left[
    \nabla_{s_{\ell}}\log p_{\param}(\vy|\vx)
    \nabla_{s_{\ell}}\log p_{\param}(\vy|\vx)^\T
    \right]\,,
    \\
    \mA_{\ell-1}
    &:=
    E\left[
    \va_{\ell-1}
    \va_{\ell-1}^\T
    \right]\,,
\end{split}
\end{equation}
and $\mG_{\ell}\in\R^{d_{\ell}\times d_{\ell}},\mA_{\ell-1}\in\R^{d_{\ell-1}\times d_{\ell-1}}$ \cite{martens2015}. 
From this definition, we can consider that K-FAC is based on an assumption 
that the input to the layer and the gradient w.r.t. the layer output are
statistically independent.

\subsubsection{K-FAC for convolutional layers}
In a convolutional (Conv) layer of a feed-forward DNN,
the output $\gS_{\ell}\in\R^{c_\ell\times h_\ell \times w_\ell}$ is calculated as
\begin{equation}
\begin{split}
    \mM_{\gA_{\ell-1}}
    &\leftarrow
    {\rm im2col}(\gA_{\ell-1})
    \in\R^{c_{\ell-1}k_{\ell}^2\times h_{\ell}w_{\ell}}\,,
    \\
    \mM_{\gS_{\ell}}
    &\leftarrow
    \mW_{\ell}\mM_{\gA_{\ell-1}}
    \in\R^{c_{\ell}\times h_{\ell}w_{\ell}}\,,
    \\
    \gS_{\ell}
    &\leftarrow
    {\rm reshape}(\mM_{\gS_{\ell}})
    \in\R^{c_{\ell}\times h_{\ell}\times w_{\ell}}
    \,,
\end{split}
\end{equation}
where $\gA_{\ell-1}\in\R^{c_{\ell-1}\times h_{\ell-1} \times w_{\ell-1}}$
is the input to this layer,
and $\mW_{\ell}\in\R^{c_{\ell}\times c_{\ell-1}k_{\ell}^2}$ is the weight matrix 
(the bias is ignored for simplicity).
$c_{\ell},c_{\ell-1}$ are the number of output, input channels, respectively,
and $k_{\ell}$ is the kernel size (assuming square kernels for simplicity).
The Kronecker factors for this Conv layer are defined as
\begin{equation}
\label{eq:kfac_conv}
\begin{split}
    \mG_{\ell}
    &:=
    E\left[
    \nabla_{\mM_{\gS_{\ell}}}\log p_{\param}(\vy|\vx)
    \nabla_{\mM_{\gS_{\ell}}}\log p_{\param}(\vy|\vx)^\T
    \right]\,,
    \\
    \mA_{\ell-1}
    &:=
    \frac{1}{h_\ell w_\ell}
    E\left[
    \mM_{\gA_{\ell-1}}
    \mM_{\gA_{\ell-1}}^\T
    \right]\,,
\end{split}
\end{equation}
and $\mG_{\ell}\in\R^{c_\ell\times c_\ell},\mA_{\ell-1}\in\R^{c_{\ell-1}k_\ell^2\times c_{\ell-1}k_\ell^2}$ \cite{grosse2016}.

\subsubsection{Inverting Kronecker-factored FIM}
By the property of the Kronecker product and 
the \textit{Tikhonov damping technique} used in \cite{martens2015},
the inverse of $\hat{\mF}_\ell+\lambda \mI$ is approximated by 
the Kronecker product of the inverse of 
each Kronecker factor
\begin{equation}
\begin{split}
    \label{eq:inv_fim}
    \left({\hat{\mF}}_{\ell}+\lambda \mI\right)^{-1}
    &=
    \left(
        {\mG}_{\ell}\otimes{\mA}_{\ell-1}+\lambda \mI
    \right)^{-1}
    \\
    &\approx
    \left(
    {\mG}_{\ell}
    +\frac{1}{\pi_{\ell}}\sqrt{\lambda}\mI
    \right)^{-1}
    \otimes
    \left(
    {\mA}_{\ell-1}
    +\pi_{\ell}\sqrt{\lambda}\mI
    \right)^{-1}
    \,,
\end{split}
\end{equation}
where $\pi_{\ell}^2$ is the average eigenvalue of $\mA_{\ell-1}$ 
divided by the average eigenvalue of $\mG_{\ell}$.
$\pi>0$ because both $\mG_{\ell}$ and $\mA_{\ell-1}$ are 
positive-semidefinite matrices as defined above.



\section{Practical Natural Gradient}
\label{sec:practical_ngd}
K-FAC for FC layer (\ref{eq:kfac_fc}) and Conv layer (\ref{eq:kfac_conv})
enables us to realize NGD in training deep ConvNets \cite{grosse2016,ba2017}.
For deep and wide neural architectures 
with a huge number of learnable parameters, 
however,
due to the extra computation for the FIM, 
even NGD with K-FAC has considerable overhead compared to SGD.
In this section, we introduce practical techniques to accelerate NGD
for such huge neural architectures.
Using our techniques, we are able to reduce the overhead of NGD
to almost a negligible amount as shown in Section~\ref{sec:experiments}.

\begin{algorithm}[t]
  \caption{\label{alg:stale_ngd}Natural gradient with the stale statistics.}
  \DontPrintSemicolon
  \SetKwInOut{Input}{input}
  \SetKwInOut{Output}{output}
  \Input{set of the statistics $\gS$ (damped inverses)}
  \Input{initial parameters $\param$}
  \Output{parameters $\param$}
  \BlankLine
  $t \gets 1$\;
  \ForEach{$X\in \gS$} {
    $t_X \gets 1$\;
  }
  \While{not converge}{
    \ForEach{$X\in \gS$} {
      \uIf{$t == t_X$}{
        $\textrm{refresh statistics}~X$\;
        $\Delta, \Delta_{-1} \gets {\rm Algorithm}~\ref{alg:interval}(X, X_{-1}, X_{-2}, \Delta, \Delta_{-1})$\;
        $t_X \gets t_X + \Delta$\;
        $X_{-1}, X_{-2} \gets X, X_{-1}$\;
      }
    }
    update $\param$ by natural gradient (\ref{eq:ngd}) using $\gS$\;
    $t \gets t+1$\;
  }
  \Return $\param$\;
\end{algorithm}

\begin{algorithm}[t]
  \caption{\label{alg:interval}
    Estimate the \textit{acceptable} interval until next refresh based on the staleness of statistics.
  }
  \DontPrintSemicolon
  \SetKwInOut{Input}{input}
  \SetKwInOut{Output}{output}
  \Input{current statistics $X$}
  \Input{last statistics $X_{-1}$}
  \Input{statistics before the last $X_{-2}$}
  \Input{last interval $\Delta_{-1}$}
  \Input{interval before the last $\Delta_{-2}$}
  \Output{next interval, last interval $\Delta,\Delta_{-1}$}
  \BlankLine
  \uIf{$X$ is not similar to $X_{-1}$}{
    $\Delta \gets \max\left\{ 1,\, \lfloor \Delta_{-1} / 2 \rfloor \right\}$\;
  }
  \uElseIf{$X$ is not similar to $X_{-2}$}{
    $\Delta \gets \Delta_{-1}$\;
  }
  \Else{
    $\Delta \gets \Delta_{-1} + \Delta_{-2}$\;
  }
  \Return $\Delta$, $\Delta_{-1}$\;
\end{algorithm}

\subsection{Fast Estimation with Empirical Fisher}
Instead of using an estimation by 
a single Monte Carlo sampling defined in Eq.~(\ref{eq:fisher_1mc})
($\hat{\mF}_{\ell,{\rm 1mc}}$),
we adopt the {\it empirical Fisher}\cite{martens2015}
to estimate the FIM $\mF_{\ell}$: 
\begin{equation}
    \mF_{\ell}
    \approx
    \hat{\mF}_{\ell,{\rm emp}}
    :=
    E\left[
    \nabla_{\vw_{\ell}}\log p_{\param}(\vt|\vx)
    \nabla_{\vw_{\ell}}\log p_{\param}(\vt|\vx)^\T
    \right]\,.
\end{equation}
We implemented an efficient $\hat{\mF}_{\ell,{\rm emp}}$ computation in the
    Chainer framework \cite{tokui2019} that allows us 
to compute $\hat{\mF}_{\ell,{\rm emp}}$ 
during the forward-pass and the backward-pass for the loss $\mathcal{L}(\param)$\footnote{The same empirical Fisher computation can be implemented on PyTorch.}.
Therefore, we \emph{do not need an extra backward-pass
to compute $\hat{\mF}_{\ell,{\rm emp}}$}, 
while \emph{an extra backward-pass is necessary for computing $\hat{\mF}_{\ell,{\rm 1mc}}$}. 
This difference is critical especially 
for a deeper network which takes longer time for a backward-pass. 

Although it is insisted that 
$\hat{\mF}_{\ell,{\rm emp}}$ is not a proper approximation of the FIM, 
and $\hat{\mF}_{\ell,{\rm 1mc}}$ is better estimation 
in the literature \cite{thomas2019,kunstner2019},
we do not see any difference in the convergence behavior nor the final model accuracy 
between NGD with $\hat{\mF}_{\ell,{\rm emp}}$ 
and that with $\hat{\mF}_{\ell,{\rm 1mc}}$
in training deep ConvNets for ImageNet classification
as shown in Section~\ref{sec:experiments}.  

\begin{figure}[t]
    \centering
    \includegraphics[width=0.8\columnwidth]{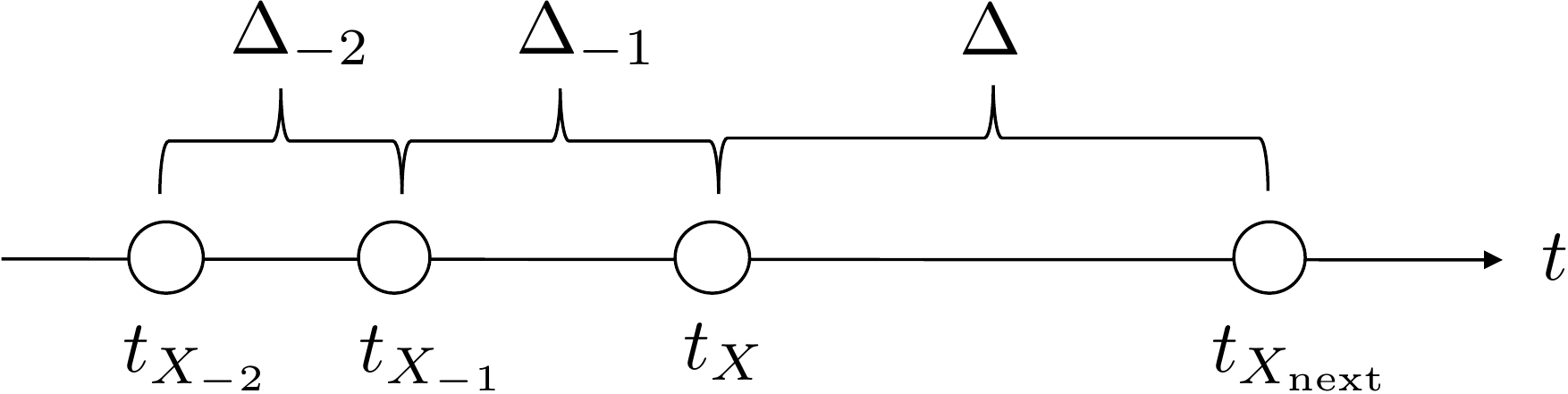}
    \caption{Estimate the interval (steps) $\Delta$ until the next step $t_{X_{\rm 
    next}}$ to refresh the statistics $X$.}
    \label{fig:interval}
\end{figure}

\subsection{Practical FIM Estimation for BatchNorm Layers}
It is often the case that a deep ConvNet has 
Conv layers that are followed by BatchNorm layers \cite{ioffe2015}.
The $\ell$-th BatchNorm layer after the $(\ell-1)$-th Conv layer 
has scale $\vgamma_\ell\in\R^{c_{\ell-1}}$ and 
bias $\vbeta_\ell\in\R^{c_{\ell-1}}$ to be applied to the normalized features.
When we see these parameters as learnable parameters, we can define
\begin{equation}
    \vw_\ell
    :=
    \left(
    \begin{array}{c c c c c}
    \gamma_{\ell,1} &
    \beta_{\ell,1} &
    \cdots &
    \gamma_{\ell,c_{\ell-1}} &
    \beta_{\ell,c_{\ell-1}}
    \end{array}
    \right)^\T
    \in\R^{2c_{\ell-1}}\,,
\end{equation}
where $\gamma_{\ell,i}$ and $\beta_{\ell,i}$ are the $i$-th element of
$\vgamma_\ell$ and $\vbeta_\ell$, respectively ($i=1,\dots,c_{\ell-1}$). 
For this BatchNorm layer,
the FIM $\mF_{\ell}$ (\ref{eq:fisher_1mc}) is a $2c_{\ell-1}\times 2c_{\ell-1}$ matrix,
and the computation cost of inverting this matrix can not be ignored
when $c_{\ell-1}$ (the number of output channels of the previous Conv layer)
is large
(e.g.~$c_{\rm out}=1024$ for a Conv layer in ResNet-50 \cite{he2016a}).

\subsubsection{Unit-wise Natural Gradient}
We approximate this FIM 
by applying unit-wise natural gradient \cite{amari2019} 
to the learnable parameters of BatchNorm layers (Figure~\ref{fig:practical_fim}).
A "unit" in a neural network is a collection of input/output nodes connected to each other.
The "unit-wise" natural gradient only takes into account the correlation of the parameters in the same node.
Hence, for a BatchNorm layer, we only consider the correlation between $\gamma_{\ell_c}$ and $\beta_{\ell,c}$ of the same channel $c$: 
\begin{equation}
\begin{split}
    \mF_{\ell}
    &\approx 
    \hat{\mF}_{\ell}
    \\
    &=
    \mF_{\ell,\rm unit\,BN}
    \\
    &:=
    {\rm diag}
    \left(
    \mF_{\ell}^{(1)}
    \dots
    \mF_{\ell}^{(i)}
    \dots
    \mF_{\ell}^{(c_{\ell-1})}
    \right)
    \in\R^{2c_{\ell-1}\times 2c_{\ell-1}}\,,
\end{split}
\end{equation}
where
\begin{equation}
    \mF_{\ell}^{(i)}
    =
    E\left[
    \begin{array}{c c}
    {\nabla_{\vgamma_{\ell}}^{(i)}}^2
    &
    \nabla_{\vgamma_{\ell}}^{(i)}
    \nabla_{\vbeta_{\ell}}^{(i)}
    \\
    \nabla_{\vbeta_{\ell}}^{(i)}
    \nabla_{\vgamma_{\ell}}^{(i)}
    &
    {\nabla_{\vbeta_{\ell}}^{(i)}}^2
    \end{array}
    \right]
    \in\R^{2\times 2}\,.
\end{equation}
$\nabla_{\vgamma_{\ell}}^{(i)}$, $\nabla_{\vbeta_{\ell}}^{(i)}$ are the $i$-th element of
$\nabla_{\vgamma_\ell}\log p_{\param}(\vy|\vx)$, $\nabla_{\vbeta_\ell}\log p_{\param}(\vy|\vx)$, respectively. 
The number of the elements to be computed and communicated
is significantly reduced from $4c_{\ell-1}^2$ to $4c_{\ell-1}$,
and we can get the inverse $\left(\mF_{\ell,{\rm unit~BN}}+\lambda \mI\right)^{-1}$ 
with little computation cost using the inverse matrix formula:
\begin{equation}
    \left[
    \begin{array}{c c}
    a & b\\
    c & d
    \end{array}
    \right]^{-1}
    =
    \frac{1}{ad-bc}
    \left[
    \begin{array}{c c}
    d & -b\\
    -c & a
    \end{array}
    \right]\,.
\end{equation}
We observed that the unit-wise approximation on $\mF_\ell$ of BatchNorm
does not affect the accuracy of a deep ConvNet on ImageNet classification
as shown in Section~\ref{sec:experiments}.

\subsection{Natural Gradient with Stale Statistics}
To achieve even faster training with NGD, it is critical to
utilize stale statistics in order to avoid re-computing the matrices
$\mA_{\ell-1},\mG_{\ell}$ and $\mF_{\ell,{\rm unitBN}}$ (Figure~\ref{fig:practical_fim}) at every step.   
Previous work on K-FAC \cite{martens2015} used a simple strategy 
where they refresh the Kronecker factors only once in 20 steps. 
However, as observed in \cite{osawa2019a}, the statistics rapidly fluctuates
at the beginning of training, and this simple strategy causes serious defects to the convergence. 
It was also observed that the degree of fluctuation of the statistics depends on
the mini-batch size, the layer, and the type of the statistics
(\eg,~statistics with a larger mini-batch fluctuates less than that with a smaller mini-batch).
Although the previous strategy \cite{osawa2019a} to reduce the frequency worked without
any degradation of the accuracy, 
it requires the prior observation on the fluctuation of the statistics,
and its effectiveness on training time has not been well studied.

\subsubsection{Adaptive Frequency To Refresh Statistics}
We propose an improved strategy 
which adaptively determines the frequency to refresh each statistics 
based on its staleness during the training. 
Our strategy is shown in Algorithm~\ref{alg:stale_ngd}.
We calculate the timing (step) to refresh each statistics 
based on the \textit{acceptable} interval (steps) estimated in
Algorithm~\ref{alg:interval} (Figure~\ref{fig:interval}).
In Algorithm~\ref{alg:interval},
matrix $A$ is considered to be \textit{similar to }matrix $B$ when
$\|A - B\|_F/\|B\|_F<\alpha$, 
where $\|\cdot\|_F$ is Frobenius norm,
and $\alpha>0$ is the threshold\footnote{$\alpha=0.1$ for all the experiments in this work.}.
We tuned $\alpha$ by running training for a few epochs to check if the threshold is not too large (preserves the same convergence). 
And it aims to find the \textit{acceptable} interval $\Delta$
where the statistics $X$ calculated at step $t=t_X+\Delta$ is \textit{similar to}
that calculated at step $t=t_X$.
With this strategy, we can estimate the \textit{acceptable} interval
for each statistics and skip the computation efficiently  
--- it keeps almost the same training time as the original, while reducing the cost for 
constructing and inverting 
$\mA_{\ell-1},\mG_{\ell}$ and $\mF_{\ell,{\rm unitBN}}$
as much as possible.
This significantly reduces the overhead of NGD.
We observe the effectiveness of our approach
in Section~\ref{sec:experiments}.

\begin{figure*}[!t]
    \centering
    \includegraphics[width=0.95\textwidth]{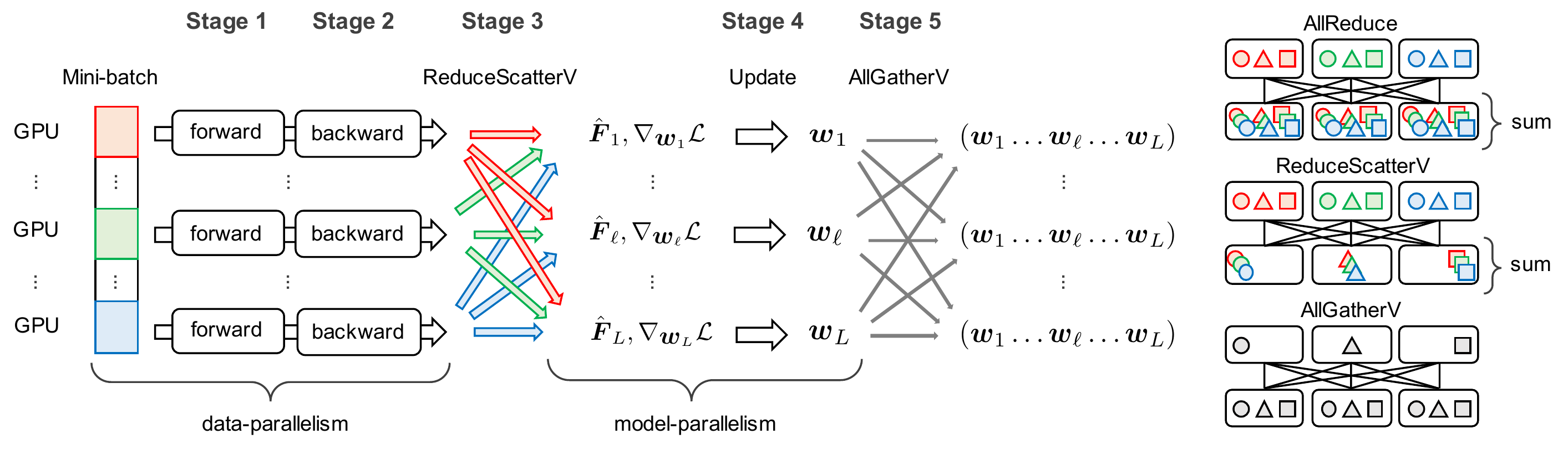}
    \caption{(Left) Overview of our distributed natural gradient descent (a single step of training). 
    (Right) Illustrations of \texttt{AllReduce}, \texttt{ReduceScatterV}, and \texttt{AllGatherV} collective. 
    Different colors correspond to data (and its communication) from different data sources.
    }
    \label{fig:dist_kfac}
\end{figure*}

\section{Scalable Natural Gradient}
\label{sec:distkfac}
Based on the practical estimation of the FIM proposed in the previous section, 
we designed a distributed parallelization scheme among multiple GPUs
so that the overhead of NGD compare to SGD
decreases as the number of GPUs (processes) is increased.



\subsection{Distributed Natural Gradient}
\begin{algorithm}[ht]
  \caption{\label{alg:distngd}Distributed Natural Gradient}
  \DontPrintSemicolon  
  \While{not converge}{
    \tcp{Stage 1}
    \ForEach{$\ell = 1, \cdots, L$} {
      forward in $\ell$-th layer\;
      \uIf{\rm $\ell$-th layer is Conv or FC}{
        compute $\mA_{\ell - 1}$
      }
    }
    \BlankLine
    \tcp{Stage 2}
    ReduceScatterV$\left(\mA_{0:L-1}\right)$\;
    \ForEach{$\ell = L, \cdots, 1$} {
      backward in $\ell$-th layer\;
      \uIf{\rm $\ell$-th layer is Conv or FC}{
        compute $\mG_{\ell}$
      }\uElseIf{\rm $\ell$-th layer is BatchNorm}{
        compute $\mF_{\ell,{\rm unitBN}}$
      }
    }
    \BlankLine
    \tcp{Stage 3}
    ReduceScatterV$\left(\mG_{1:L}/\mF_{1:L,{\rm unitBN}}
    \mbox{ and }\nabla_{\vw_{1:L}}\mathcal{L} \right)$\;
    \BlankLine
    \tcp{Stage 4}
    \ForPar{$\ell = 1, \cdots, L$} {
        update $\vw_{\ell}$ by natural gradient (\ref{eq:ngd})
    }
    \BlankLine
    \tcp{Stage 5}
    AllGatherV$\left(\vw_{1:L}\right)$\;
  }
  \Return $\boldsymbol{\theta}=\left(\vw_1^\T,\dots,\vw_L^\T\right)^\T$\;
\end{algorithm}
Figure \ref{fig:dist_kfac} shows the overview of our design,
which shows a single step of training with our distributed NGD.
We use the term \texttt{Stage} to refer to 
each phase of computation and communication, 
which is indicated at the top of the figure.
Algorithm \ref{alg:distngd} shows the pseudo code of our distributed NGD design.

In \texttt{Stage 1}, each process (GPU) receives 
a different part of the mini-batch and performs a forward-pass 
in which the Kronecker factor $\mA_{\ell - 1}$ is computed for the received samples,
if $\ell$-th layer is a Conv layer or a FC layer.

In \texttt{Stage 2}, two procedures are performed in parallel
--- communication among all the processes and a backward-pass in each process.
Since \texttt{Stage 1} is done in a {\bf data-parallel} fashion,
each process computes the statistics only for the different parts of the mini-batch.
In order to compute these statistics for the entire mini-batch,
we need to average these statistics over all the processes.
This is performed using a \texttt{ReduceScatterV} collective communication,
which transitions our approach from data-parallelism to model-parallelism 
by reducing (taking the sum of) $\mA_{\ell-1}$ for different $\ell$ to different processes.
This collective is much more efficient than an \texttt{AllReduce}
, where $\mA_{\ell-1}$ for all $\ell$ are reduced to all the processes (Figure~\ref{fig:dist_kfac}).
While $\mA_{\ell-1}$ is communicated, 
each process also performs a backward-pass 
to get the gradient $\nabla_{\vw_{\ell}}\mathcal{L}$, 
the Kronecker factor $\mG_{\ell}$ for Conv, FC layers,
and $\mF_{\ell,{\rm unitBN}}$ for BatchNorm layer, for each $\ell$.

In \texttt{Stage 3}, 
$\mG_{\ell}$,$\mF_{\ell,{\rm unitBN}}$ and $\nabla_{\vw_{\ell}}\mathcal{L}$  
are communicated in the same way as $\mA_{\ell}$ 
by \texttt{ReduceScatterV} collective.
At this point, only a single process has 
the FIM estimation $\hat{\mF}_{\ell}$ and 
the gradient $\nabla_{\vw_{\ell}}\mathcal{L}$  
with the statistics for the entire mini-batch for the $\ell$-th layer.
In \texttt{Stage 4}, only the process that has the FIM computes the matrix inverse and 
applies the NGD update (\ref{eq:ngd}) to the weights $\vw_{\ell}$ of the $\ell$-th layer. 
Hence, these computations are performed in a {\bf model-parallel} fashion.
When the number of layers is larger than the number of processes,
multiple layers are handled by a process.

Once the weights $\vw_{\ell}$ of each $\ell$ are updated,
we synchronize the updated weights 
among all the processes by calling an \texttt{AllGatherV} (Figure~\ref{fig:dist_kfac}) collective,
and we switch back to data-parallelism.
Combining the practical estimation of the FIM proposed in the previous section,
we are able to reduce a significant amount of communication required
for the Kronecker factors $\mA_{\ell-1},\mG_{\ell}$ and $\mF_{\ell,{\rm unitBN}}$.
Therefore, the amount of communication for our distirbuted NGD is similar to distributed SGD,
where the \texttt{AllReduce} for the gradient $\nabla_{\vw_{\ell}}\mathcal{L}$ 
is implemented as a \texttt{ReduceScatter+AllGather}.

\subsection{Further acceleration}
Our data-parallel and model-parallel hybrid approach allows us to 
minimize the overhead of NGD in a distributed setting.
However, NGD still has a large overhead compared to SGD.
There are two hotspots in our distributed NGD design.
The first is the construction of the statistics 
$\mA_{\ell-1},\mG_{\ell}$, and $\mF_{\ell,{\rm unitBN}}$,
that cannot be done in a model-parallel fashion.
The second is the communication (\texttt{ReduceScatterV}) for distributing these statistics.
In this section, we discuss how we accelerate these two hotspots
to achieve even faster training time.

\medskip
\noindent
\textbf{Mixed-precision computation.}
We use the Tensor Cores in the NVIDIA Volta Architecture \footnote{https://www.nvidia.com/en-us/data-center/tensorcore/}.
This more than doubles the speed of the calculation for this part.
One might think 
that this low-precision computation affects the overall accuracy
of the training, but in our experiments 
we do not find any differences between 
training with half-precision floating point computation and 
that with full-precision floating point computation.


\medskip
\noindent
\textbf{Symmetry-aware communication.}
The statistics matrices $\mA_{\ell-1},\mG_{\ell}$, and $\mF_{\ell,{\rm unitBN}}$
are symmetric matrices.
We exploit this property to reduce the amount of communication
without loss of information.
To communicate a symmetric matrix of size $N \times N$,
we only need to send the upper triangular matrix with $N(N + 1)/2$ elements.

In addition to these two optimizations,
we also adopted the performance optimizations done by \cite{tsuji2019}: 
\begin{itemize}
    \item Explicitly use NHWC (mini-batch, height, width, and channels) format for the input/output data (tensor) of Conv layers instead of NCHW format.
    This makes cuDNN API to fully benefit from the Tensor Cores.
    \item Data I/O pipeline using the NVIDIA Data Loading Library (DALI) \footnote{https://developer.nvidia.com/DALI}.
    \item Hierarchical \texttt{AllReduce} collective proposed by Ueno \etal \cite{ueno2019}, which alleviates the latency of the \texttt{ring-AllReduce} communication among a large number of GPUs.
    \item Half-precision communication for \texttt{AllGatherV} collective.
\end{itemize}
\section{Training for ImageNet Classification}
\label{sec:training}
The behavior of NGD on large models and datasets has not been studied in depth.
Also, there are very few studies that use NGD (K-FAC) for large mini-batches (over 4K)
using distributed parallelism at scale \cite{ba2017}.
Contrary to SGD, where the hyperparameters have been optimized by many practitioners even for large mini-batches,
there is very little insight on how to tune hyperparameters for NGD.
In this section, we have explored some methods, which we call training schemes,
to achieve higher accuracy in our experiments.
In this section, we show those training schemes 
in our large mini-batch training with NGD for ImageNet classification.

\subsection{Data augmentation}
\label{subsec:data}
To achieve good generalization performance while keeping 
the benefit of the fast convergence that comes from NGD,
we adopt the data augmentation techniques commonly used for
training networks with large mini-batch sizes.
We resize all the images in ImageNet to $256\times256$ ignoring the aspect ratio of original images and compute the mean value of the upper left portion ($224\times224$) of the resized images.
When reading an image, we randomly crop a $224\times224$ image from it,
randomly flip it horizontally, subtract the mean value, and scale every pixel to $[0, 1]$.

\medskip
\noindent
{\bf Running mixup.} We extend {\it mixup} \cite{zhang2018b} to increase its regularization effect.
We synthesize virtual training samples from raw samples and virtual samples from the previous step
(while the original {\it mixup} method synthesizes new samples only from the raw samples):
\begin{align}
    \tilde{\vx}^{(t)}&=\lambda\cdot
    \vx^{(t)}+(1-\lambda)\cdot\tilde{\vx}^{(t-1)}\,,
    \\
    \tilde{\vt}^{(t)}&=\lambda\cdot
    \vt^{(t)}+(1-\lambda)\cdot\tilde{\vt}^{(t-1)}\,.
\end{align}
$\vx^{(t)},\vt^{(t)}$ is a raw input and label (one-hot vector),
and $\tilde{\vx}^{(t)},\tilde{\vt}^{(t)}$ is a virtual input and label for $t$ th step.
$\lambda$ is sampled from the Beta distribution with the beta function
\begin{equation}
B(\alpha,\beta)
=\int_0^1t^{\alpha-1}(1-t)^{\beta-1}\,\mathrm{d}t\,.
\end{equation}
where we set $\alpha=\beta=\alpha_{\rm mixup}$.

\medskip
\noindent
{\bf Random erasing with zero value.} We also implemented {\it Random Erasing} \cite{zhong2017}.
We set elements within the erasing region of each input to zero instead of a random value as used in the original method.
We set the erasing probability $p=0.5$, the erasing area ratio $S_e\in[0.02,0.25]$, and the erasing aspect ratio $r_e\in[0.3,1]$.
We randomly switch the size of the erasing area from $(H_e,W_e)$ to $(W_e,H_e)$.

\subsection{Learning rate and momentum}
\label{subsec:lr}
The learning rate used for all of our experiments is scheduled by {\it polynomial decay}.
The learning rate $\eta^{(e)}$ for $e$ th epoch is determined as follows:
\begin{equation}
    \eta^{(e)}
    =
    \eta^{(0)}
    \cdot
    \left(
        1-
        \frac{e-e_{\rm start}}
        {e_{\rm end} - e_{\rm start}}
    \right)^{p_{\rm decay}}.
\end{equation}
$\eta^{(0)}$ is the initial learning rate and $e_{\rm start},e_{\rm end}$ is the epoch when the decay starts and ends.
The decay rate $p_{\rm decay}$ guides the speed of the learning rate decay.

We use the momentum method for NGD updates.
Because the learning rate decays rapidly in the final stage of the training with the polynomial decay, the current update can become smaller than the previous update.
We adjust the momentum rate $m^{(e)}$ for $e$ th epoch so that the ratio between $m^{(e)}$ and $\eta^{(e)}$ is fixed throughout the training: 
\begin{equation}
     m^{(e)}=\frac{m^{(0)}}{\eta^{(0)}}\cdot\eta^{(e)}\,,
\end{equation}
where $m^{(0)}$ is the initial momentum rate.
The weights are updated as follows:
\begin{equation}
    \vw^{(t+1)}_\ell
    \leftarrow
    \vw^{(t)}_\ell
    -\eta^{(e)}
    {\left(\hat{\mF}_{\ell}^{(t)}+\lambda \mI\right)}^{-1}\nabla_{\vw_\ell}\gL^{(t)}
    +m^{(e)}\vv^{(t)}\,,
    \label{eq:paramupdate}
\end{equation}
where $\vv^{(t)}=\vw^{(t)}_\ell-\vw^{(t-1)}_\ell$.

\subsection{Weights rescaling}
\label{subsec:rescale}
To prevent the scale of weights from becoming too large,
we adopt the {\it Normalizing Weights} \cite{vanlaarhoven2017} technique
to the $\vw_\ell$ of FC and Conv layers.
We rescale the $\vw_\ell$ to have a norm $\sqrt{2\cdot d_{\rm out}}$ after (\ref{eq:paramupdate}):
\begin{align}
    \vw_\ell^{(t+1)}
    &\leftarrow
    \sqrt{2\cdot d_{\rm out}}\cdot
    \frac{
        \vw_\ell^{(t+1)}
    }
    {
        \|\vw_\ell^{(t+1)}\|+\epsilon
    }\,.
\end{align}
where we use $\epsilon=1\cdot 10^{-9}$ to stabilize the computation.
$d_{\rm out}$ is the output dimension or channels of the layer.

\section{Experiments}
\label{sec:experiments}

We train ResNet-50 \cite{he2016a} for ImageNet \cite{deng2009} in all of our experiments.
We use the same hyperparameters for the same mini-batch size.
The hyperparameters for our results are shown in Table~\ref{tab:settings}. 
We implement all of our methods using Chainer \cite{tokui2019}.
Our Chainer extenstion is available at https://github.com/tyohei/chainerkfac.
We initialize the weights by the \texttt{HeNormal} initializer of Chainer \footnote{https://docs.chainer.org/en/stable/reference/generated/\\chainer.initializers.HeNormal.html} with the default parameters.

\subsection{Experiment Environment}
We conduct all experiments on the ABCI (AI Bridging Cloud Infrastructure)
\footnote{https://abci.ai/}
operated by the National Institute of Advanced Industrial Science and Technology (AIST) in Japan.
ABCI has 1088 nodes with four NVIDIA Tesla V100 GPUs per node.
Due to the additional memory required by NGD, all of our experiments use a mini-batch size of 32 per GPU.
We were only given a 24 hour window to use the full machine 
so we had to tune the hyperparameters on a smaller number of nodes 
while mimicking the global mini-batch size of the full node run.
For large mini-batch size experiments which cannot be executed directly
(BS=65K, 131K requires 2048, 4096 GPUs, respectively)
,we used an accumulation method to mimic the behavior 
by accumulating the statistics $\mA_{\ell-1},\mG_{\ell},\mF_{\ell,{\rm unitBN}}$, and $\nabla_{\vw_{\ell}}\mathcal{L}$ over multiple steps.

\begin{table*}[t]
    \centering
    \caption{The hyperparameters of the training with large mini-batch size (BS) used for our schemes in Section \ref{sec:training} and top-1 single-crop validation accuracy of ResNet-50 for ImageNet.
    \texttt{reduction} and \texttt{speedup} correspond to 
    the reduction rate of the communication amount 
    and the speedup comes from that, respectively, 
    for \texttt{emp+unitBN+stale} compare to \texttt{emp+unitBN} in Figure~\ref{fig:secpiter}.
    }
    \label{tab:settings}
    \begin{tabular}{c c c c c c c c | c c c c c}
        \toprule
        BS & $\alpha_{\rm mixup}$ & $p_{\rm decay}$ & $e_{\rm start}$ & $e_{\rm end}$ & $\eta^{(0)}$ & $m^{(0)}$ & $\lambda$ & \# steps & top-1 accuracy & reduction $\downarrow$ & speedup $\uparrow$\\
        \midrule
        4K & $0.4$ & $11.0$ & $1$ & $53$ & $8.18\cdot10^{-3}$ & $0.997$ & $2.5\cdot10^{-4}$ & 10,948 & 75.2~\% $\to$ 74.8~\% & 23.6~\% & $\times~1.33$ \\
        8K & $0.4$ & $8.0$ & $1$ & $53.5$ & $1.25\cdot10^{-2}$ & $0.993$ & $2.5\cdot10^{-4}$ & 5,434 & 75.5~\% $\to$ 75.3~\% & 15.1~\% & $\times~1.32$ \\
        16K & $0.4$ & $8.0$ & $1$ & $53.5$ & $2.5\cdot10^{-2}$ & $0.985$ & $2.5\cdot10^{-4}$ & 2,737 & 75.3~\% $\to$ 75.2~\% & {\bf 5.4}~\% & $\times~{\bf 1.68}$\\
        32K & $0.6$ & $3.5$ & $1.5$ & $49.5$ & $3.0\cdot10^{-2}$ & $0.97$ & $2.0\cdot10^{-4}$ & 1,760 & 75.6~\% $\to$ 75.4~\% & 7.8~\% & $\times~1.40$\\
        65K & $0.6$ & $2.9$ & $2$ & $64.5$ & $4.0\cdot10^{-2}$ & $0.95$ & $1.5\cdot10^{-4}$ & 1,173 & 75.6~\% & n/a & n/a \\
        131K & $1.0$ & $2.9$ & $3$ & $100$ & $7.0\cdot10^{-2}$ & $0.93$ & $1.0\cdot10^{-4}$ & 873 & 74.9~\% & n/a & n/a \\
        \bottomrule
    \end{tabular}
\end{table*}

\subsection{Extremely Large Mini-batch Training}
\label{subsec:training}
We trained ResNet-50 for ImageNet classification task
with extremely large mini-batch size BS=\{4,096, 8,192, 16,384, 32,768, 65,536, 131,072\} 
and achieved a competitive top-1 validation accuracy ($\geq74.9\%$)
compared to highly-tuned SGD training. 
The summary of the training is shown in Table~\ref{tab:compare}.
For BS=\{4K, 8K, 16K, 32K, 65K\}, the training converges in much less than 90 epochs,
which is the usual number of epochs required by SGD-based training of ImageNet \cite{akiba2017,goyal2017,jia2018,mikami2018,you2017}. 
For BS=\{4K,8K,16K\}, the required epochs to reach higher than 75\% top-1 validation accuracy is 35 epochs.
Even for a relatively large mini-batch size of BS=32K, 
NGD still converges in 45 epochs, half the number of epochs compared to SGD.
Note that the calculation time is still decreasing while the number of epochs
is less than double when we double the mini-batch size,
assuming that doubling the mini-batch corresponds to doubling the number of GPUs (and halving the execution time).
When we increase the mini-batch size to BS=\{32K,65K,131K\}, 
we see a significantly small number of steps it takes to converge.
At BS=32K and 65K, it takes 1760 steps (45 epochs) and 1173 steps (60 epochs), respectively.
At BS=131K, there are less than 10 iterations per epoch since the dataset size of ImageNet is 1,281,167, and it only takes 873 steps to converge (90 epochs).
None of the SGD-based training of ImageNet have sustained the top-1 validation accuracy at this mini-batch size.
Furthermore, this is the first work that uses NGD for the training with extremely large mini-batch size
BS=\{16K,32K,65K,131K\} and achieves a competitive top-1 validation accuracy.

\subsection{Scalability}
\begin{figure}[!t]
    \centering
    \includegraphics[width={\linewidth}]{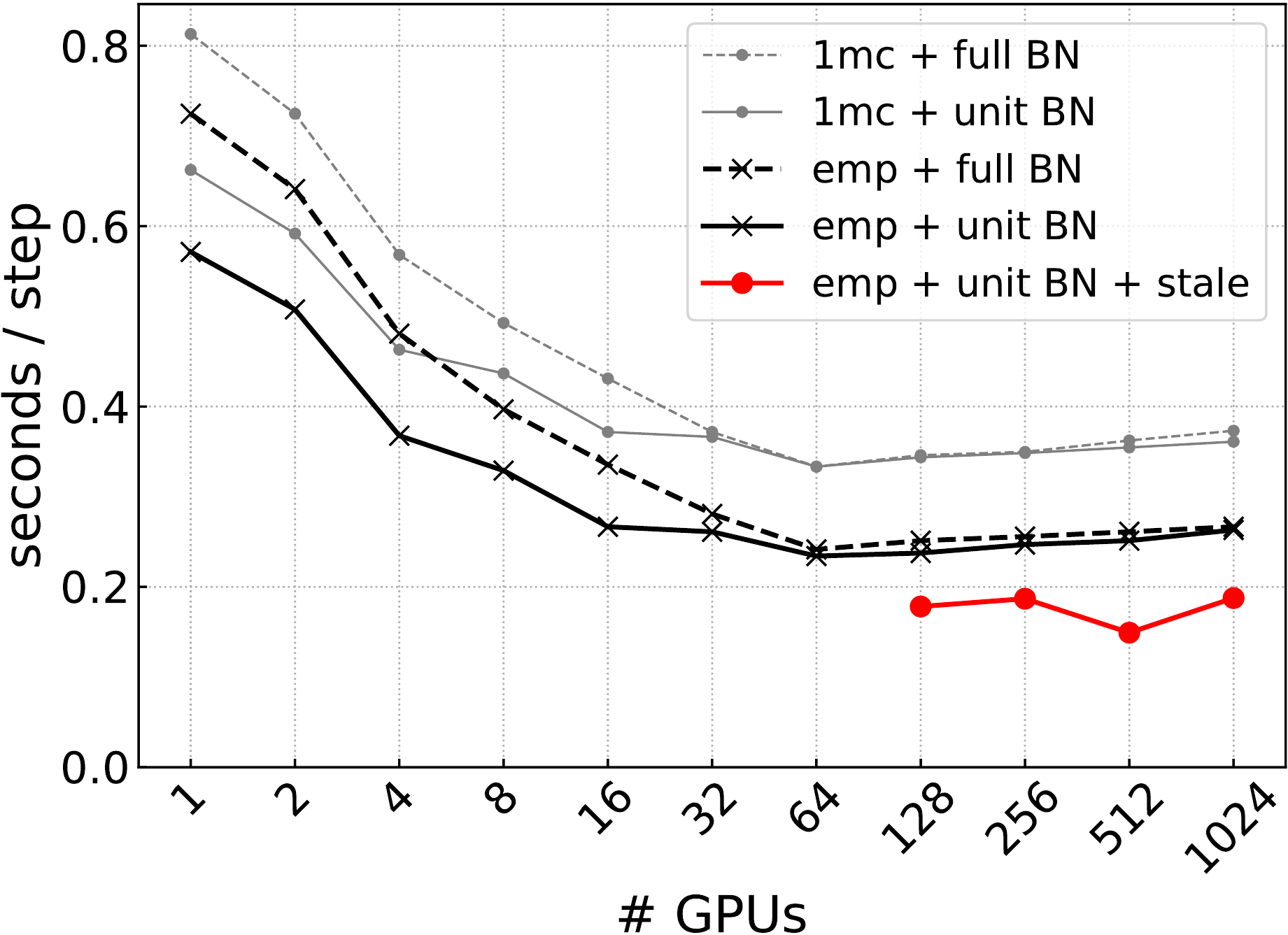}
    \caption{Time per step for trainig ResNet-50 (107 layers in total) 
    on ImageNet with our scalable and practical NGD. Each GPU processes 32 images.
    \texttt{1mc} and \texttt{emp} correspond to 
    NGD with $\hat{\mF}_{\ell,{\rm 1mc}}$ and 
    that with $\hat{\mF}_{\ell,{\rm emp}}$, respectively.
    \texttt{fullBN} and \texttt{unitBN} correspond to 
    NGD and unit-wise NGD on BatchNorm parameters, respectively.
    \texttt{stale} corresponds to NGD with the stale statistics. 
    }
    \label{fig:secpiter}
\end{figure}
We measure the scalability of our distributed NGD implementation 
for training ResNet-50 on ImageNet.
Figure~\ref{fig:secpiter} shows the time for one step
with different number of GPUs and different techniques proposed in Section~\ref{sec:practical_ngd}.
Note that since we fix the number of images to be processed per GPU (=32), 
doubling the number of GPUs means doubling the total number of images (mini-batch size) to be processed in a step (\eg, 32K images are processed with 1024 GPUs in a step).
In a distributed training with multiple GPUs,
it is considered ideal if this plot shows a flat line parallel to the x-axis,
that is, the time per step is independent of the number of GPUs,
and the number of images processed in a certain time increases linearly with the number of GPUs.
From 1 GPU to 64 GPUs, however, we observed a \textit{superlinear} scaling.
For example, the time per step with 64 GPUs is 300\% faster than 
that with 1 GPU for \texttt{emp+fullBN}.
This is the consequence of our model-parallel design 
since ResNet-50 has 107 layers in total when all the Conv, FC,
and BatchNorm layers are accounted for.
With more than 128 GPUs, we observe slight performance
degradation due to the communication overhead comes from
\texttt{ReduceScatterV} and \texttt{AllGatherV} collective.
Yet for \texttt{emp+unitBN+stale}, we see almost the ideal scaling
from 128 GPUs to 1024 GPUs.
Moreover, with 512 GPUs, which corresponds to BS=16K, 
we see a \textit{superlinear} scaling, again.
We discuss this in the next sub-section.

\subsection{Effectiveness of Practical Natural Gradient}
\begin{figure*}[t]
    \centering
    \includegraphics[width=\textwidth]{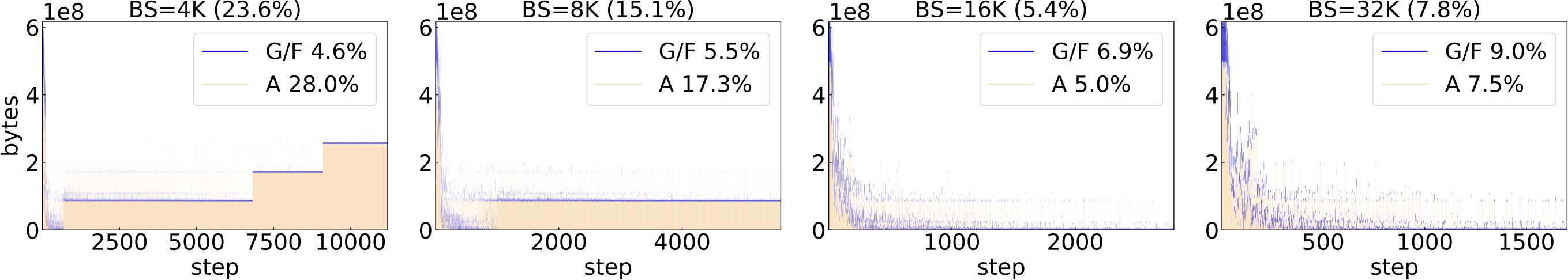}
    \caption{The communication amount (bytes) for the statistics
    ($\mA_{\ell-1},\mG_{\ell},\mF_{\ell,{\rm unitBN}}$)
    in each step in training ResNet-50 on ImageNet with BS=\{4K,8K,16K,32K\}
    (stacked graph --- the amount for \texttt{G/F} is stacked on the amount for \texttt{A}).
    \texttt{A} and \texttt{G/F} correspond to the communication amount for
    $\mA_{\ell-1}$ and $\mG_{\ell}/\mF_{\ell,{\rm unitBN}}$, respectively. 
    The reduction rate (smaller is better) of the communication amount 
    for all the statistics throughout the training is shown with the percentage (\%).
    }
    \label{fig:stale}
\end{figure*}
\label{subsec:practical_ngd}
We examine the effectiveness of our practical NGD approaches 
proposed in Section~\ref{sec:practical_ngd}
for training ResNet-50 on ImageNet with extreamely large mini-batch.
We show that our practical techniques makes the overhead of NGD
close to a negligible amount and improves training time significantly.
The summary of the training time is shown in Table~\ref{tab:compare} and Figure~\ref{fig:accuracy_iter}.

\medskip
\noindent
{\bf Natural Gradient by Empirical Fisher.}
We compare the time and model accuracy in training by 
NGD with empirical Fisher and
that with a Fisher estimation by a single Monte Carlo sampling
($\hat{\mF}_{\ell,{\rm emp}}$ vs $\hat{\mF}_{\ell,{\rm 1mc}}$) .
In Figure~\ref{fig:secpiter}, the time per a step by each training is labeled as 
\texttt{emp} and \texttt{1mc}, respectively. 
Due to the extra backward-pass required for constructing $\hat{\mF}_{\ell,{\rm 1mc}}$,
\texttt{1mc} is slower than \texttt{emp} at any number of GPUs.
We do not see any difference in the convergence behavior (accuracy vs steps)
and the final accuracy for training ResNet-50 on ImageNet with BS=\{4K,8K,16K,32K,65K,131K\}.
Note that we used the same hyperparameters tuned for \texttt{emp} for each BS
(shown in Table~\ref{tab:settings}) 
for the limitation of computational resource to tune for \texttt{1mc}.

\medskip
\noindent
{\bf Unit-Wise Natural Gradient.}
We also compare training with natural gradient and that with unit-wise natural gradient on BatchNorm parameters ($\mF_{\ell}$ vs $\mF_{\ell,{\rm unitBN}}$) .
In Figure~\ref{fig:secpiter}, the time per step for each method is labeled as 
\texttt{fullBN} and \texttt{unitBN}, respectively. 
From 1 GPU to 16 GPUs, we can see that 
\texttt{unitBN} effectively accelerates the time per step compare to \texttt{fullBN}.
For more than 32 GPUs, we can see only a slight improvements since 
inverting statistics ($\mA_{\ell-1},\mG_{\ell}$ and $\mF_{\ell}$) for all the layers are already
distributed among enough number of processes,
and inverting $\mF_{\ell}$ is no longer a bottleneck of processing a step. 
We do not see any difference in the convergence behavior (accuracy vs steps)
and the final accuracy for training ResNet-50 on ImageNet with BS=\{4K,8K,16K,32K,65K,131K\}.

\medskip
\noindent
{\bf Natural Gradient with Stale Statistics.}
We apply our adaptive frequency strategy described in Section~\ref{sec:practical_ngd}
to training ResNet-50 on ImageNet with BS=\{4K,8K,16K,32K\}.
In Figure~\ref{fig:secpiter}, the time per a step with all the practical techniques 
is labeled as \texttt{emp+unitBN+stale}.
The model accuracy before and after applying this technique, 
reduction rate (smaller is better) of the communication volume for the statistics, 
and speedup (\texttt{emp+unitBN} vs \texttt{emp+unitBN+stale}) 
is shown in Table~\ref{tab:settings}.
The communication amount (bytes) in the \texttt{ReduceScatterV} collective 
in each step during a training 
and the reduction rate are plotted in Figure~\ref{fig:stale}.
With BS=16K,32K, we can reduce the communication amount for the statistics
($\mA_{\ell-1},\mG_{\ell}$ and $\mF_{\ell}$) to 5.4\%,7.8\%, respectively. 
We might be able to attribute this significant reduction rate to the fact that
the statistics with larger BS (16K,32K) is more stable than that with smaller BS (4K,8K).
Note that though we show the reduction rate of the amount of communication,
this rate is also applicable to estimate the reduction rate of the amount of computation
for the statistics, and the cost for inverting them is also removed.
With these improvements on NGD,
we see almost an ideal scaling from 128 GPUs to 1024 GPUs, which corresponds to
BS=4K to 32K.

\subsection{Training ResNet-50 on ImageNet with NGD in 5.5 minutes}
Finally, we combine all the practical techniques 
--- empirical Fisher, unit-wise NGD and NGD with stale statistics.
Using 1024 NVIDIA Tesla V100, we achieve 75.4 \% top-1 accuracy with ResNet-50 for ImageNet
in 5.5 minutes (1760 steps = 45 epochs, including a validation after each epoch).
We used the same hyperparameters shown in Table~\ref{tab:settings}.
The training time and the validation accuracy are competitive with the results reported
by related work that use SGD for training (the comparison is shown in Table~\ref{tab:compare}).
We refer to our training method as \textit{Scarable and Practical NGD} (SP-NGD).

\section{Discussion and Future Work}
\label{sec:conclusion}
In this work, we proposed a \textit{Scalable and Practical Natural Gradient Descent} (SP-NGD),
a framework which combines i) a large-scale distributed computational design with data and model hybrid parallelism for the Natural Gradient Descent (NGD) \cite{amari1998} 
and ii) practical Fisher information estimation techniques including 
Kronecker-Factored Approximate Curvature (K-FAC) \cite{martens2015}, 
that alleviates the computational overhead of NGD over SGD.
Using our SP-NGD framework, 
we showed the advantages of the NGD over first-order stochastic gradient descent (SGD) 
for training ResNet-50 on ImageNet classification with extremely large mini-batches.
We introduced several schemes for the training using the NGD with mini-batch sizes up to 131,072
and achieved over 75\% top-1 accuracy in much fewer number of steps compared to
the existing work using the SGD with large mini-batch.
Contrary to prior claims that models trained with second-order methods do not generalize as well as the SGD,
we were able to show that this is not at all the case, even for extremely large mini-batches.
Our SP-NGD framework allowed us to train on 1024 GPUs
and achieved 75.4\% in 5.5 minutes.
This is the first work which observes the relationship between the FIM of ResNet-50
and its training on large mini-batches ranging from 4K to 131K.
The advantage that we have in designing better optimizers by taking this approach
is that we are starting from the most mathematically rigorous form,
and every improvement that we make is a systematic design decision based on observation of the FIM.
Even if we end up having similar performance to the best known first-order methods,
at least we will have a better understanding of why it works by starting from second-order methods including NGD.

\medskip
\noindent
{\bf More accurate and efficient FIM estimation.}
We showed that NGD using the empirical Fisher matrix \cite{martens2015} (\texttt{emp})
is much faster than that with an estimation using a single Monte Carlo (\texttt{1mc}), 
which is widely used by related work on the approximate natural gradient.
Although it is stated that \texttt{emp} is not a good approximation of the NGD
in the literature \cite{kunstner2019, thomas2019}, we observed the same convergence behavior as
\texttt{1mc} for training ResNet-50 on ImageNet.
We might be able to attribute this result to the fact 
that \texttt{emp} is a good enough approximation 
to keep the behavior of the \textit{true} NGD 
or that even \texttt{1mc} is not a good approximation. 
To know whether these hypotheses are correct or not 
and to examine the actual value of the true NGD,
we need a more accurate and effcient estimation of the NGD 
with less computational cost. 

\medskip
\noindent
{\bf Towards Bayesian deep learning.}
NGD has been applied to Bayesian deep learning for estimating 
the posterior distribution of the network parameters.
For example, K-FAC \cite{martens2015} has been applied to Bayesian deep learning to realize \textit{Noisy Natural Gradient} \cite{zhang2018a},
and our distributed NGD has been applied to that at ImageNet scale \cite{osawa2019}.
We similarly expect that our SP-NGD framework will accelerate Bayesian deep learning research using natural gradient methods.

\ifCLASSOPTIONcompsoc
  \section*{Acknowledgments}
\else
  \section*{Acknowledgment}
\fi
Computational resource of AI Bridging Cloud Infrastructure (ABCI) was awarded by "ABCI Grand Challenge" Program,
National Institute of Advanced Industrial Science and Technology (AIST).
This work was supported by JSPS KAKENHI Grant Number JP18H03248 and JP19J13477.
This work is supported by JST CREST Grant Number JPMJCR19F5, Japan.
(Part of) This work is conducted as research activities of AIST - Tokyo Tech Real World Big-Data Computation
Open Innovation Laboratory (RWBC-OIL).
This work is supported by "Joint Usage/Research Center for Interdisciplinary Large-scale
Information Infrastructures" in Japan (Project ID: jh180012-NAHI).
This research used computational resources of the HPCI system provided by (the names of the HPCI System Providers) through the HPCI System Research Project (Project ID:hp190122).
We would like to thank Yaroslav Bulatov (South Park Commons) for helpful comments on the manuscript.

\ifCLASSOPTIONcaptionsoff
  \newpage
\fi
\bibliographystyle{IEEEtran}
\bibliography{biblio}

\begin{IEEEbiography}[{\includegraphics[width=1in,height=1.25in,clip,keepaspectratio]{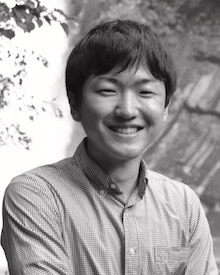}}]{Kazuki Osawa}
received his BS and MS from Tokyo Institute of Technology in 2016 and 2018, respectively.
He is currently a PhD candidate at Tokyo Institute of Technology and a Research Fellow of Japan Society for the Promotion of Science (JSPS).
His research interests include optimization, approximate Bayesian inference, and distributed computing for deep learning. 
He is a student member of IEEE.
\end{IEEEbiography}

\begin{IEEEbiography}[{\includegraphics[width=1in,height=1.25in,clip,keepaspectratio]{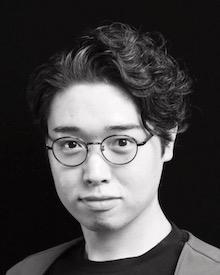}}]{Yohei Tsuji}
received his BS and MS from Tokyo Institute of Technology in 2017 and 2019, respectively.
He is currently a PhD student at Tokyo Institute of Technology.
His research interests include high performance computing for machine learning, probabilistic programming.
\end{IEEEbiography}

\begin{IEEEbiography}[{\includegraphics[width=1in,height=1.25in,clip,keepaspectratio]{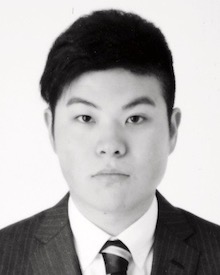}}]{Yuichiro Ueno}
received his BS from Tokyo Institute of Technology in 2019. He is currently a master course student at Tokyo Institute of Technology. His research interests include a range of high-performance computing, such as GPU computing and networking, and its application to deep learning. He is a student member of ACM.
\end{IEEEbiography}

\begin{IEEEbiography}[{\includegraphics[width=1in,height=1.25in,clip,keepaspectratio]{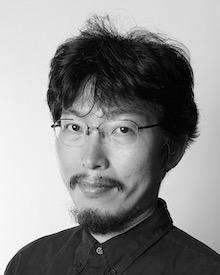}}]{Akira Naruse}
is a senior developer technology engineer at NVIDIA. He holds the MS degree in computer science from Nagoya University. Prior to joining NVIDIA, he was a research engineer at Fujitsu Laboratory and was involved in various high performance computing projects. His main interest is the performance analysis and optimization of scientific computing and deep learning applications on very large systems.
\end{IEEEbiography}

\begin{IEEEbiography}[{\includegraphics[width=1in,height=1.25in,clip,keepaspectratio]{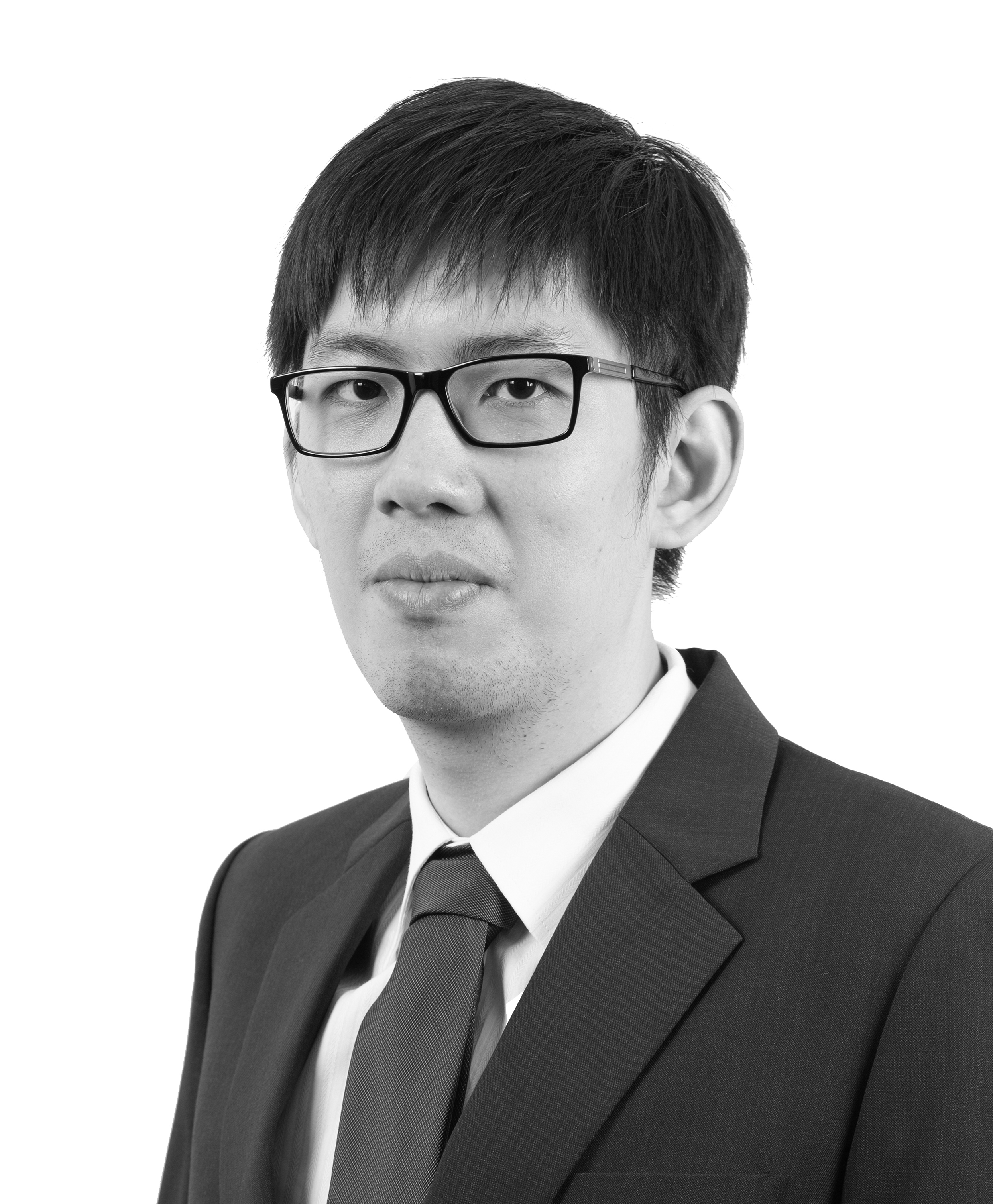}}]{Chuan-Sheng Foo}
is a Scientist at the Institute for Infocomm Research, A*STAR. He received his BS, MS and PhD from Stanford University. His research focuses on developing deep learning algorithms that can learn from less labeled data, inspired by applications in healthcare and medicine. 
\end{IEEEbiography}

\begin{IEEEbiography}[{\includegraphics[width=1in,height=1.25in,clip,keepaspectratio]{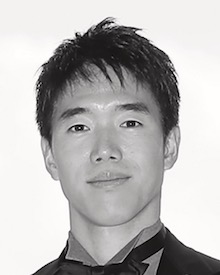}}]{Rio Yokota}
received his BS, MS, and PhD from Keio University in 2003, 2005, and 2009, respectively. 
He is currently an Associate Professor at GSIC, Tokyo Institute of Technology. 
His research interests range from high performance computing, hierarchical low-rank approximation methods, and scalable deep learning. 
He was part of the team that won the Gordon Bell prize for price/performance in 2009.
\end{IEEEbiography}

\end{document}